\documentclass[runningheads]{llncs}
\usepackage{graphicx}
\usepackage{tabularx}
\usepackage[misc,geometry]{ifsym} 
\usepackage{subfigure}
\begin{document}
\title{Boosting Object Recognition in Point Clouds by Saliency Detection}
%
%
\author{Marlon Marcon\inst{1,2,3}(\Letter) \and
Riccardo Spezialetti\inst{3} \and
Samuele Salti\inst{3} \and
Luciano Silva\inst{2} \and
Luigi Di Stefano\inst{3}}
\authorrunning{M. Marcon et al.}
%
\institute{Federal University of Technology - Paran\'{a}, Dois Vizinhos, Brazil\\ \email{marlonmarcon@utfpr.edu.br}\and
Federal University of Paran\'{a}, Curitiba, Brazil\\
\email{luciano@inf.ufpr.br} \and
University of Bologna, Bologna, Italy\\
\email{\{riccardo.spezialetti, samuele.salti, luigi.distefano\}@unibo.it}}
\maketitle              
\begin{abstract}

Object recognition in 3D point clouds is a challenging task, mainly when time is an important factor to deal with, such as in industrial applications. Local descriptors are an amenable choice whenever the 6 DoF pose of recognized objects should also be estimated. However, the  pipeline for this kind of descriptors is highly time-consuming. In this work, we propose an update to the traditional pipeline, by adding a preliminary filtering stage   referred to as saliency boost. We perform tests on a standard object recognition benchmark by considering four keypoint detectors and four local descriptors, in order to compare time and recognition performance between the traditional pipeline and the boosted one. Results on time show that the boosted pipeline could turn out up to 5 times faster, with the recognition rate improving in most of the cases and exhibiting only a slight decrease in the others. These results suggest that the boosted pipeline can speed-up processing time substantially with limited impacts or even benefits in recognition accuracy.
\end{abstract}

\section{Introduction} 

Application of computer vision techniques aimed at object recognition is gathering increasing attention in industrial applications. Among others, prominent applications in this space include robot picking in assembly lines and surface inspection. To address these tasks, the vision  system must estimate the 6 DoF (degrees-of-freedom) pose of the sought objects, which calls for a 3D object recognition approach. Moreover, in industrial settings robustness, accuracy as well as run-time performance are particularly important.  

Reliance on RGB-D sensors providing both depth and color information is conducive to 3D object recognition. Yet, typical nuisances to be dealt with in 3D object recognition applications include clutter, occlusions and the significant degree of noise which affects most  RGB-D cameras. Many studies, such as \cite{Johnson1999,Guo2015IJCV}, have investigated on these problems and highlighted how local 3D descriptors can effectively withstand clutter, occlusions and noise in 3D object recognition. 

The local descriptors pipeline for 3D object recognition is however quite slow. Indeed, RGB-D cameras generate a high amount of data (over 30MB/s) and, as this may hinder performance in  embedded and real-time applications, sampling strategies are needed. To reduce processing time, keypoint extraction techniques are widely used. In addition, some solutions propose to assign higher priority to specific image areas, like, for example, in the foveation technique \cite{Gomes2013}. Another approach, inspired by human perception and widely explored for 2D image segmentation, consists in saliency detection, which identifies the most prominent points within an image \cite{Aytekin2018}. Unlike the foveation, which processes arbitrary regions, the use of  saliency allows for highlighting image regions that are known to be more important.

This work proposes a solution to improve the performance of the standard local descriptors pipeline for 3D object recognition from point clouds. The idea consists in  adding a preliminary step, referred to as Saliency Boost, which filters the point clouds using a saliency mask in order to reduce the number of processed points and consequently the whole processing time. Besides, by selecting only salient regions, our approach may  yield a reduction in the number of false positives, thereby often also enhancing object recognition accuracy. 

\section{Related Works}

3D object recognition systems based on local descriptors typically deploy two stages, one carried out offline and the other online, referred to as training and testing, respectively. The training stage builds the database of objects, storing their features for later use. In the testing stage, then,  features are extracted from scene images. Given a scene, the typical pipeline, depicted in Figure \ref{fig:proposed-pipeline} and described, e.g. in \cite{Chen2007}, consists of the following steps  1) Keypoints extraction; 2) Local descriptors calculation; 3) Matching; 4) Grouping correspondences; and 5) Absolute orientation estimation. The first two, described in more details below, are those which really differentiate the various approaches and impact performance most directly. 

\subsection{Keypoints Extraction}

This step concerns selecting some surface points, either from  images or point clouds. According to \cite{Tombari2012},  keypoint extraction must reduce data dimensionality without losing discriminative capacity. In this work, we explore techniques which work in 3D, as Uniform sampling and Intrinsic Shape Signatures (ISS), and 2D alike, i.e. SIFT and FAST.

 Uniform Sampling downsamples the point cloud segmenting it in voxels  based on a certain leaf size, and selects as keypoint each  nearest neighbor point to a voxel centroid \cite{PCL}.  ISS \cite{Zhong2009} selects keypoints based on a local surface saliency criterion, so as to extract 3D points that exhibit a large surface variation in their neighbourhood. 

 The keypoint detector proposed in SIFT \cite{sift1999} is arguably the prominent proposal for RGB images. It is based on the detection of blob-like and high contrast local features amenable to compute highly distinctive features and similarity invariant image descriptors. The FAST keypoint extractor is a 2D corner detector based on a machine learning approach, which is  widely used in real-time computer vision applications due to its remarkable computational efficiency.

\subsection{Local Descriptors}
\label{sec:local-features}

A local 3D descriptor processes the neighborhood of a keypoint to produce a feature vector discriminative with respect  to clutter and robust to noise. Many descriptors have been proposed in recent years and several works, e.g. \cite{Guo2015IJCV}, have investigated on their relative merits and limitations.  In this work, we explore both descriptors which process only depth information, such as Signatures of Histograms of OrienTations (SHOT)\cite{Salti2014} and  Fast Point Feature Histogram (FPFH)\cite{Rusu2009}, as well as depth and color, like Point Feature Histogram for Color (PFHRGB) \cite{Rusu2008} and Color SHOT (CSHOT) \cite{Salti2014}.

Introduced by \cite{Salti2014}, SHOT describes a keypoint based on spatial and geometric information. To calculate the descriptor, first 3D Local Reference Frame (LRF) around the keypoint is established. Then, a canonical spherical grid is  divided into 32 segments. Each segment results in a histogram that describes the angles between the normals at the keypoint and the normal at the neighboring points. The authors also proposed a variation to work with color at the points, called CSHOT. The color value is encoded according to the  CIELab color space and added to the angular information deployed in SHOT. This descriptor is known to yield better results than SHOT when applied to colored point clouds.

PFHRGB \cite{Rusu2008} is based on the Point Feature Histogram (PFH) and stores geometrical information by analyzing the angular variation of the normal between each pair of combination in a set composed by the keypoint and all its k-neighbors.  PFHRGB works on RGB and stores also the color ratio between the keypoint and its neighbors, increasing its efficiency on RGB-D data \cite{Alexandre2012}. In order to speed-up the descriptor calculation, \cite{Rusu2009} proposed a simplified solution, called FPFH, which considers only the differences between the keypoint and its k-neighbors. Also, an influence weight is stored, resulting in a descriptor which can be calculated faster while maintaining its discriminative capacity.

\subsection{Saliency Detection}

Salient object detection is a topic inspired by human perception, which affirms that the human being tends to select visual information based on attention mechanisms in the brain \cite{Kastner2000}. Its objective is to emphasize regions of interest in a scene \cite{Aytekin2018}. Many applications benefit from the use of saliency, such as object tracking and recognition, image retrieval, restoring and segmentation.

The majority of recent works perform saliency detection using either RGB  \cite{Aytekin2018,Hou2019} or RGB-D \cite{Chen2018,Li2018} images and are based on Deep Learning algorithms.

\section{Proposed Approach}

\begin{figure}[htb]
\centering
\includegraphics[width=\textwidth]{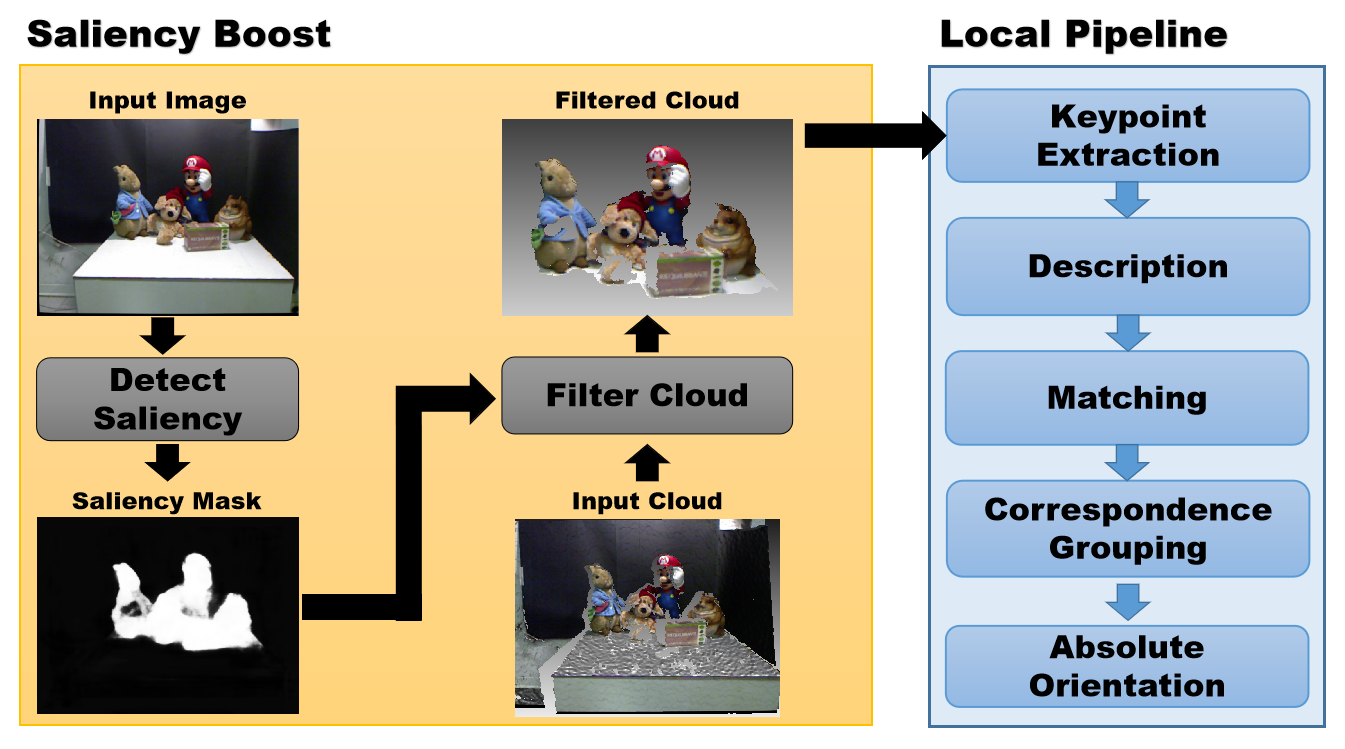}
\caption{Local descriptor pipeline with saliency boost.}
\label{fig:proposed-pipeline}
\end{figure}

We present a way to improve significantly the time performance and also the memory efficiency of the standard pipeline described above, by adding an additional step to the original pipeline. We refer to this step as \textit{saliency boost}.
It leverages the RGB scene image by detecting salient regions within it, which are then used to filter the point cloud and to execute the local descriptors' pipeline only on salient regions. In particular, we use the saliency mask to reduce the search space for 3D keypoints by letting them run on the part of the point cloud which corresponds to the salient regions of the image. To project saliency information from the 2D domain of the RGB image to the point cloud we leverage the registration information provided by RGB-D cameras. Figure \ref{fig:proposed-pipeline} presents a graphical overview of the approach. In the case of 2D keypoint detectors, instead, we run them on the full RGB image and we then filter out keypoints not belonging to the salient regions: we do not filter the image before the keypoint extraction step because 2D detectors exploit also pixels from the background to define blobs and edges/corners to detect keypoints. In the 3D case, instead, points from the background are usually far away and outside the sphere used to define the keypoint neighborhood, so it is possible to filter them before without affecting the detector performance.

Our approach is not dependent from a specific saliency detection technique. In this work, we choose the DSS algorithm \cite{Hou2019}, and we detect salient areas by running the trained model provided by the authors. 

\section{Experimental Results}
\subsection{Dataset}

\begin{figure}[!htb]
\centering
\includegraphics[width=\textwidth]{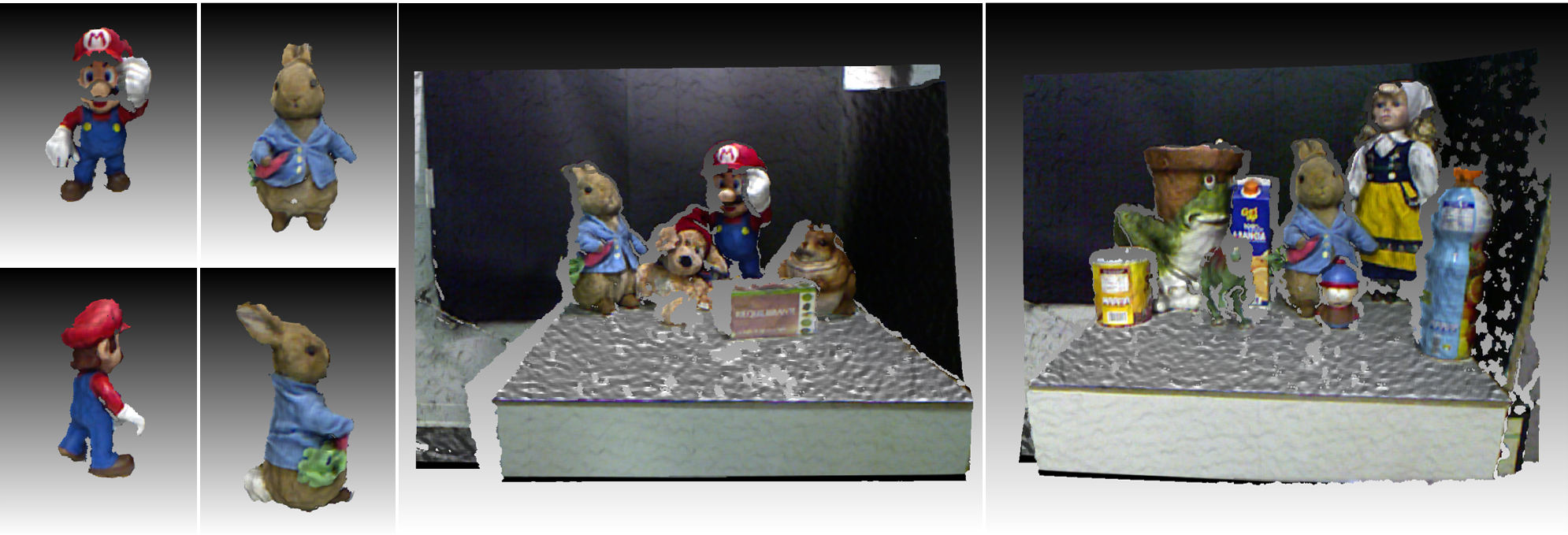}
\caption{Examples of models and scenes from Kinect dataset \cite{Salti2014}. Models of Mario and Rabbit (leftmost figures), and scenes (rightmost figures).}
\label{fig:kinect-dataset}
\end{figure}

The experiments were performed on the Kinect dataset from the University of Bologna, presented in \cite{Salti2014}. This dataset has sixteen scenes, and six models with pose annotation. Each model is represented as a set of 2.5D views from different angles and has from thirteen to twenty samples. Figure \ref{fig:kinect-dataset} depicts some examples of models and scenes in this dataset.

\subsection{Local Descriptors Pipeline}

In the local feature pipeline for object recognition, the choice of the keypoint extraction and description methods is key, and depends on the applications, the kind of 3D representation available and their resolution, the sensor noise, etc... In order to evaluate the performance of the proposed approach in an application-agnostic scenario, we test combinations of several descriptors and detectors. The selected descriptors are: SHOT and CSHOT (Color SHOT) \cite{Salti2014}, FPFH \cite{Rusu2009} and PFHRGB \cite{Rusu2008}. The keypoint detectors working on 3D data are Uniform sampling (with leaf sizes ranging from 2 to 5 cm with step of 1 cm) and ISS \cite{Zhong2009}, while on images we test SIFT \cite{sift1999} and FAST \cite{fast2006}, run on the RGB image and projected on the point cloud, as discussed. 

The matching step is performed by nearest neighbor (NN) search implemented by the FLANN library, integrated in the Point Cloud Library (PCL) \cite{PCL}. A KdTree is built for each view of each model in the database and each keypoint on the scene is matched to only one point of one view of one model in the database by selecting the closest descriptor among views and models. After this process, all matches pointing to a view of a model are processed by the Geometric Consistency Grouping algorithm \cite{Chen2007}, which selects all the subsets of geometrically consistent matches between the view and the scene, and estimates the aligning transformation. The transformation obtained from the largest correspondence group among all the views of an object is considered the best estimation of the aligning transformation for that object. If an object fails to have a geometrically consistent subset with at least 3 matches among all its views, it is estimated as being not present in the considered scene.

\subsection{Evaluation Protocol}

In order to evaluate the performance of the proposed object detection pipeline, the correctness of predictions both of object presence and pose are evaluated. We adopt the Intersection over Union (IoU) metric (Equation \ref{eq:IoU}), also known as the Jaccard index, and defined as the ratio between the intersection and the union of the estimated bounding box ($BB_{Est}$) and the ground truth bounding box($BB_{GT}$). 

\begin{equation} \label{eq:IoU}
IoU = \frac{BB_{GT} \cap BB_{Est}}{BB_{GT} \cup BB_{Est}}
\end{equation}

A detection is evaluated as correct if its IoU with the ground truth is greater than 0.25, as in \cite{Song2016}. Given detections and ground truth boxes, we calculate precision and recall (Equations \ref{eq:precision} and \ref{eq:recall}) by considering a correct estimation as True Positive ($ TP $), i.e. $IoU \geq 0.25$, an estimation of an absent object as False Positive ($FP$), and misdetections or detections with $IoU < 0.25$ as False Negative ($ FN $).

\begin{equation} \label{eq:precision}
precision = \frac{TP}{(TP + FP)}
\end{equation}

\begin{equation} \label{eq:recall}
recall = \frac{TP}{(TP + FN)}
\end{equation}

To calculate precision-recall curves (PRC), we varied the threshold on the number of geometrically consistent correspondences to  declare a detection, increasing it from the minimum value of 3 up to when no more detections are found in a scene. The area under the PRC curve (AUC) is then computed for each combination detector/descriptor and used to compare and rank the pipelines.

\subsection{Implementation Details}

Tests were performed on a Linux Ubuntu 16.04 LTS machine, using the Point Cloud Library (PCL), Version 1.8.1, OpenCV 3.4.1 and the VTK 6.2 library. For comparison purposes, all trials were performed on the same computer, equipped with an Intel Core i7-3632QM processor and 8GB of RAM. When available in PCL, the parallel version of each descriptor was used (e.g. for SHOT, CSHOT, and FPFH).

As for parameters of detectors, the ISS Non-Maxima Suppression radius was set to $0.6$ cm and the neighborhood radius to $1$ cm, while for SIFT and FAST we used the default values provided in OpenCV. As for descriptors, to estimate the normals we used the first ten neighbors of each point while the description radius was set to $5$ cm for all the considered.

\subsection{Results}

In this section, we present the results obtained in the experiments. All trials were performed on the Kinect dataset, comparing the original pipeline (blue part in Figure \ref{fig:proposed-pipeline}) with the proposed pipeline with saliency boosting. For each descriptor and each pipeline we tested seven keypoint extractors, totaling 56 trials. The scene processing time, which comprises the saliency detection (only for boosted pipeline), keypoint extraction, description, matching correspondences, clustering and pose estimation, was measured to verify the impact of the proposed modification also on processing time.

Results in terms of the number of keypoints extracted are presented in Table \ref{tab:keypoints-extracted}. The saliency filtering reduces significantly the average number of keypoins extracted by each detector: reduction using saliency boost ranges from $24.58\%$ to almost $80\%$ with an average of $56\%$.

\begin{table}[!htb]
\centering
\caption{Average number of keypoints extracted from scenes in the trials with the traditional local pipeline (LP) and boosted by saliency (Boost). The column ``\%'' represents the variation between them. Best value in bold.}\label{tab:keypoints-extracted}
\begin{tabular}{p{3cm}|r|r|r}
\hline
\textbf{Keypoints} &  \multicolumn{1}{c|}{\textbf{LP}} &  \multicolumn{1}{c|}{\textbf{Boost}} &  \multicolumn{1}{c}{\textbf{\%}} \\ \hline
FAST & 489.71 & 369.36 & -24.58 \\
ISS  & 4201.16 & 846.75 & \textbf{-79.84} \\
SIFT & 282.79 & 199.79 & -29.35 \\
US$_{0.02}$ & 4559.80 & 1457.86 & -68.03 \\
US$_{0.03}$ & 2144.07 & 731.36 & -65.89 \\
US$_{0.04}$ & 1266.00 & 446.29 & -64.75 \\
US$_{0.05}$ & 820.57 & 303.71 & -62.99 \\ \hline
\textbf{Average} & \multicolumn{1}{c}{-} & \multicolumn{1}{c}{-} & \multicolumn{1}{r}{-56.49} \\ \hline
\end{tabular}
\end{table}

The number of keypoints extracted impacts directly the running time of the pipeline, mainly by two factors: the number of descriptors that have to be computed and the time it takes to match them. The SHOT and CSHOT descriptors are calculated relatively fast but due to their length (352 and 1344 bins respectively), the matching phase is slower, accounting for 97 and 99\% of the processing time, respectively. The PFHRGB and FPFH are shorter descriptors (250 and 33 bins respectively), but the description is slower and requires 94 and 89\% of the overall time, respectively. As shown in Table \ref{tab:results-time}, the extraction of keypoints only in salient regions reduces drastically the processing time for both kinds of descriptors.

In the best case, reductions in processing time is as high as 80\%, i.e. the boosted pipeline is 5 times faster due to the proposed saliency boosting. For all the considered detector/descriptor combinations, deployment of the saliency boosting step always reduces the processing time significantly, from the 22\% obtained by FAST/SHOT to 83\% for ISS and US$_{0.05}$ with FPFH. 

\begin{table}[!htb]
\scriptsize
\centering
\caption{Average scene processing time (s) in the trials with the traditional Local Pipeline (LP) and boosted by saliency (Boost). The column ``\%'' represents the variation between them. Best value in each column in bold.}\label{tab:results-time}
\begin{tabular}{l|rrr|rrr|rrr|rrr}
\hline
 & \multicolumn{3}{c|}{\textbf{CSHOT}} & \multicolumn{3}{c|}{\textbf{SHOT}} &
\multicolumn{3}{c|}{\textbf{PFHRGB}} & \multicolumn{3}{c}{\textbf{FPFH}} \\ \hline
\textbf{Keypoints} & \multicolumn{1}{c}{\textbf{LP}} & \multicolumn{1}{c}{\textbf{Boost}} & \multicolumn{1}{c|}{\textbf{\%}} & \multicolumn{1}{c}{\textbf{LP}} & \multicolumn{1}{c}{\textbf{Boost}} & \multicolumn{1}{c|}{\textbf{\%}} & \multicolumn{1}{c}{\textbf{LP}} & \multicolumn{1}{c}{\textbf{Boost}} & \multicolumn{1}{c|}{\textbf{\%}} & \multicolumn{1}{c}{\textbf{LP}} & \multicolumn{1}{c}{\textbf{Boost}} & \multicolumn{1}{c}{\textbf{\%}} \\ \hline
FAST & 244.0 & 174.8 & -28.36 & 59.1 & 45.9 & -22.31 & 351.4 & 238.8 & -32.06 & 46.6 & 19.0 & -59.14 \\
ISS  & 226.1 & \textbf{47.7} & \textbf{-78.90} & 72.4 & \textbf{17.1} & \textbf{-76.45} & 2580.4 & 489.1 & \textbf{-81.05} & 141.9 & 24.3 & -82.92 \\
SIFT & \textbf{132.2} & 94.5 & -28.50 & \textbf{34.3} & 25.7 & -25.29 & \textbf{195.3} & 138.2 & -29.24 & \textbf{31.3} & \textbf{17.7} & -43.39 \\
US$_{0.02}$ & 2167.9 & 668.8 & -69.15 & 505.7 & 174.5 & -65.50 & 2100.9 & 455.5 & -78.32 & 150.6 & 29.6 & -80.32 \\
US$_{0.03}$ & 988.0 & 335.6 & -66.04 & 238.8 & 88.0 & -63.16 & 913.2 & 191.5 & -79.03 & 137.4 & 24.1 & -82.47 \\
US$_{0.04}$ & 583.4 & 205.2 & -64.83 & 139.7 & 54.1 & -61.29 & 506.1 & 103.5 & -79.56 & 130.9 & 22.2 & -83.08 \\
US$_{0.05}$ & 378.1 & 139.9 & -63.01 & 90.7 & 37.2 & -58.99 & 304.3 & \textbf{62.1} & -79.61 & 128.7 & 20.9 & \textbf{-83.76} \\ \hline
\textbf{Average} &  &  & -56.97 &  &  & -53.29 &  &  & -65.55 &  &  & -73.58 \\ \hline
\end{tabular}
\end{table}

Reducing processing time is only beneficial if it doesn't harm recognition and localization performance. Interestingly, deployment of the saliency boosting step very often improves AUC with respect to the traditional pipeline, as shown in Table \ref{tab:results-AUC}. In particular, for 19 of the 28 trials which included the saliency boosting step, the pipeline boosted by saliency performed better also on AUC, with massive improvements by more than $50 \%$ for PFHRGB and FPFH. Viceversa, when the AUC decreases due to the deployment of the saliency boost, it does it usually marginally, by $1$ or $2\%$, with the worst decrease in AUC being greater than $10\%$ only once, when using the SIFT detector.

\begin{table}[!htb]
\scriptsize
\centering
\caption{AUC results in the trials with the traditional Local Pipeline (LP) and boosted by saliency (Boost).  The column ``\%'' represents the variation between them. Best value in each column in bold.}\label{tab:results-AUC}
\begin{tabular}{l|rrr|rrr|rrr|rrr}
\hline
 & \multicolumn{3}{c|}{\textbf{CSHOT}} & \multicolumn{3}{c|}{\textbf{SHOT}} & \multicolumn{3}{c|}{\textbf{PFHRGB}} & \multicolumn{3}{c}{\textbf{FPFH}} \\ \hline
\textbf{Keypoints} & \multicolumn{1}{c}{\textbf{LP}} & \multicolumn{1}{c}{\textbf{Boost}} & \multicolumn{1}{c|}{\textbf{\%}} & \multicolumn{1}{c}{\textbf{LP}} & \multicolumn{1}{c}{\textbf{Boost}} & \multicolumn{1}{c|}{\textbf{\%}} & \multicolumn{1}{c}{\textbf{LP}} & \multicolumn{1}{c}{\textbf{Boost}} & \multicolumn{1}{c|}{\textbf{\%}} & \multicolumn{1}{c}{\textbf{LP}} & \multicolumn{1}{c}{\textbf{Boost}} & \multicolumn{1}{c}{\textbf{\%}} \\ \hline
FAST & 0.946 & 0.874 & -7.61 & 0.915 & 0.892 & -2.45 & 0.743 & 0.761 & 2.43 & 0.631 & 0.668 & 5.89 \\
ISS & 0.868 & 0.881 & 1.52 & 0.866 & 0.912 & \textbf{5.30} & \textbf{0.745} & \textbf{0.900} & 20.68 & 0.491 & \textbf{0.752} & \textbf{53.04} \\
SIFT & 0.864 & 0.889 & 2.83 & 0.903 & 0.820 & -9.15 & 0.472 & 0.549 & 16.41 & 0.529 & 0.476 & -10.13 \\
US$_{0.02}$ & \textbf{0.949} & \textbf{0.948} & -0.07 & \textbf{0.941} & \textbf{0.938} & -0.31 & 0.739 & 0.807 & 9.19 & \textbf{0.641} & 0.728 & 13.48 \\
US$_{0.03}$ & 0.861 & 0.905 & 5.08 & 0.875 & 0.843 & -3.58 & 0.731 & 0.814 & 11.37 & 0.488 & 0.621 & 27.26 \\
US$_{0.04}$ & 0.832 & 0.875 & 5.23 & 0.824 & 0.817 & -0.92 & 0.564 & 0.700 & 24.22 & 0.289 & 0.368 & 27.14 \\
US$_{0.05}$ & 0.582 & 0.619 & \textbf{6.19} & 0.682 & 0.644 & -5.64 & 0.373 & 0.599 & \textbf{60.76} & 0.145 & 0.162 & 11.77 \\ \hline
\textbf{Average} &  &  & 1.88 &  &  & -2.39 &  &  & 20.72 &  &  & 18.35 \\ 
\hline
\end{tabular}
\end{table}

While the AUC generally increases with the boosted pipeline, it doesn't do so on average when deployed with the SHOT descriptor. However, it does increase by $5\%$ in the very relevant case of combining SHOT with the ISS detector, the combination that delivers the fastest running time among all the tested variants (as shown in Table \ref{tab:results-time}).

\begin{figure}[!htb]
\centering
\subfigure{
\includegraphics[width=0.47\textwidth]{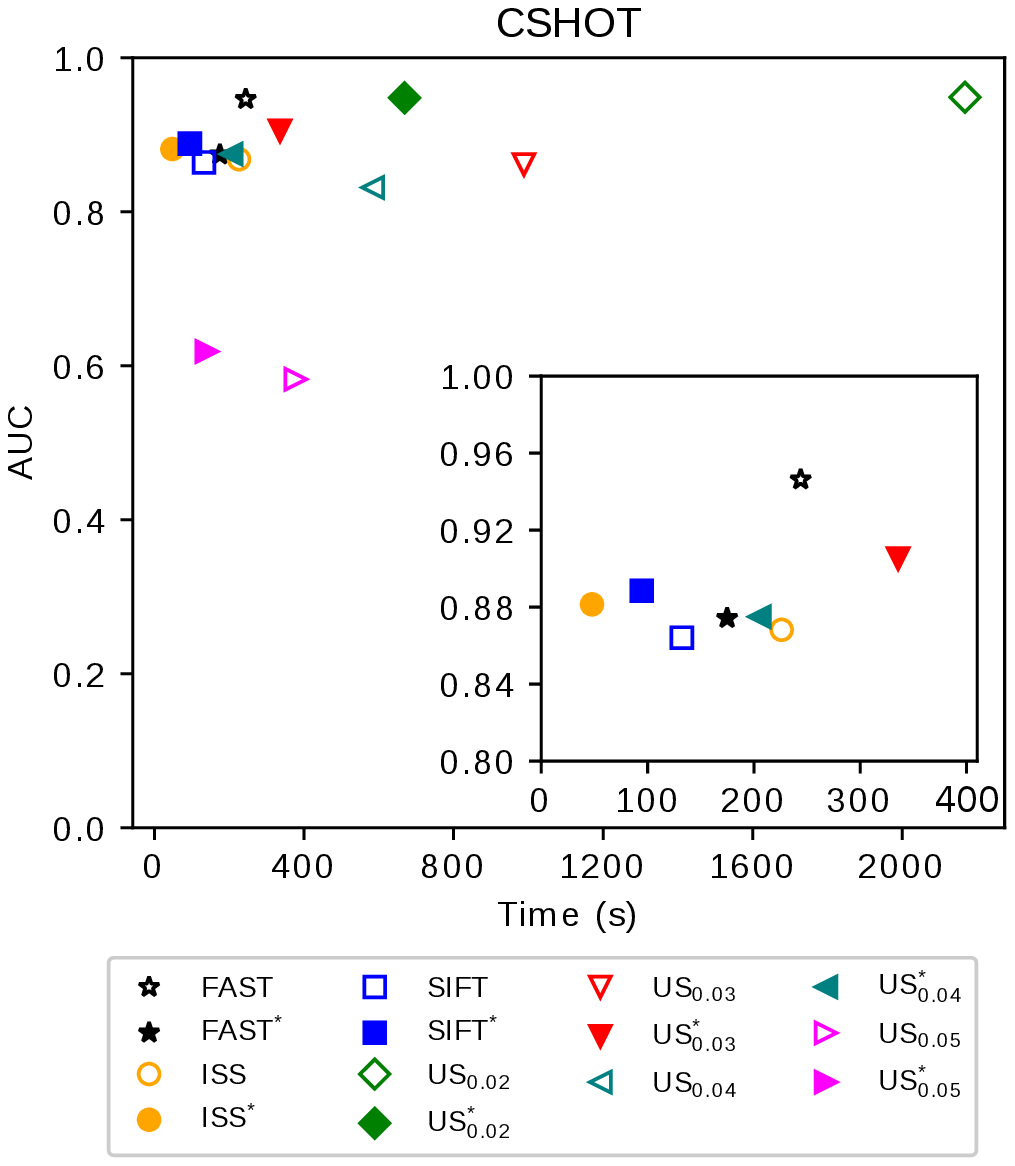}
\label{fig:cshot-results}
}
\subfigure{
\includegraphics[width=0.47\textwidth]{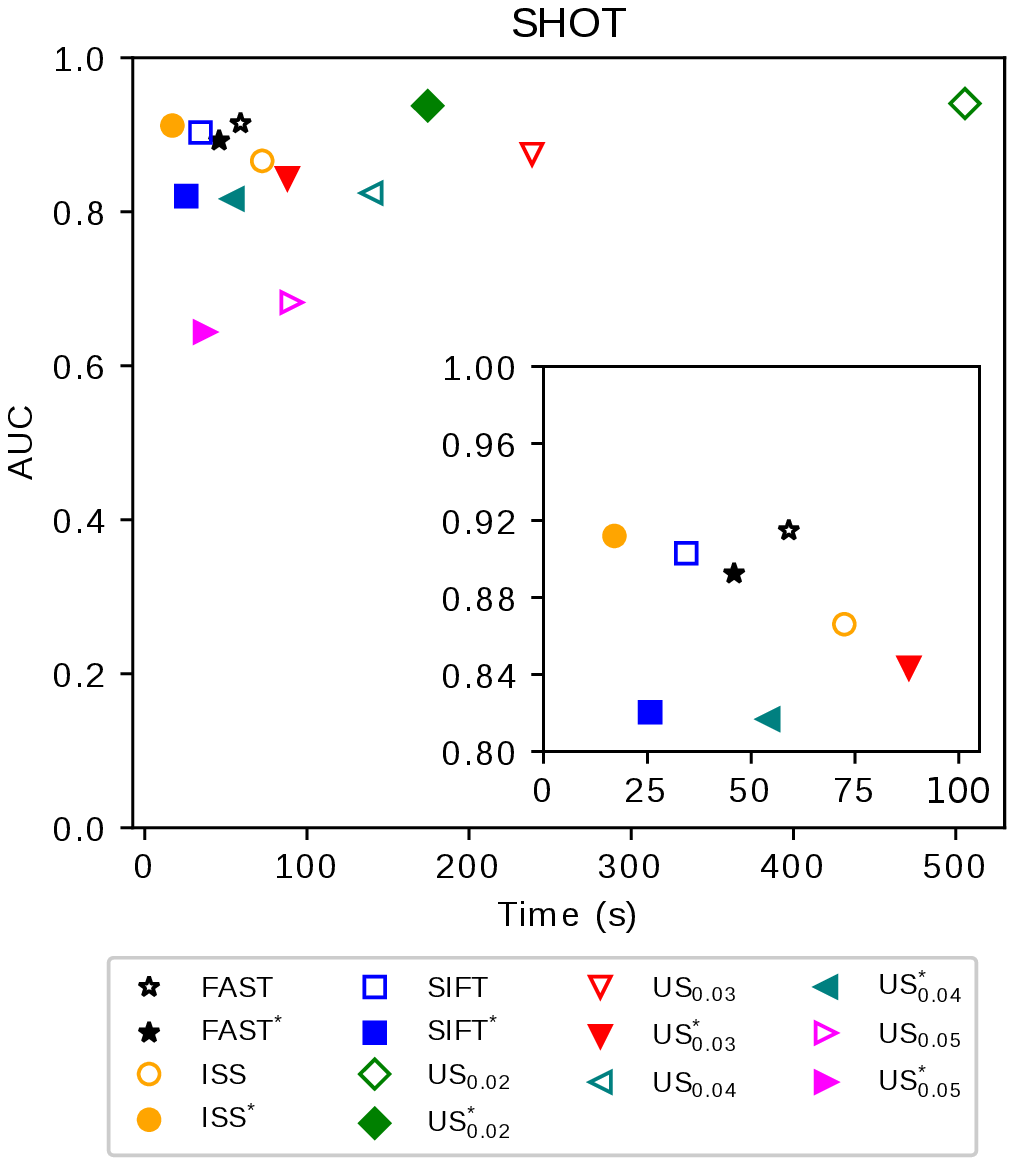}
\label{fig:shot-results}
}
\subfigure{
\includegraphics[width=0.47\textwidth]{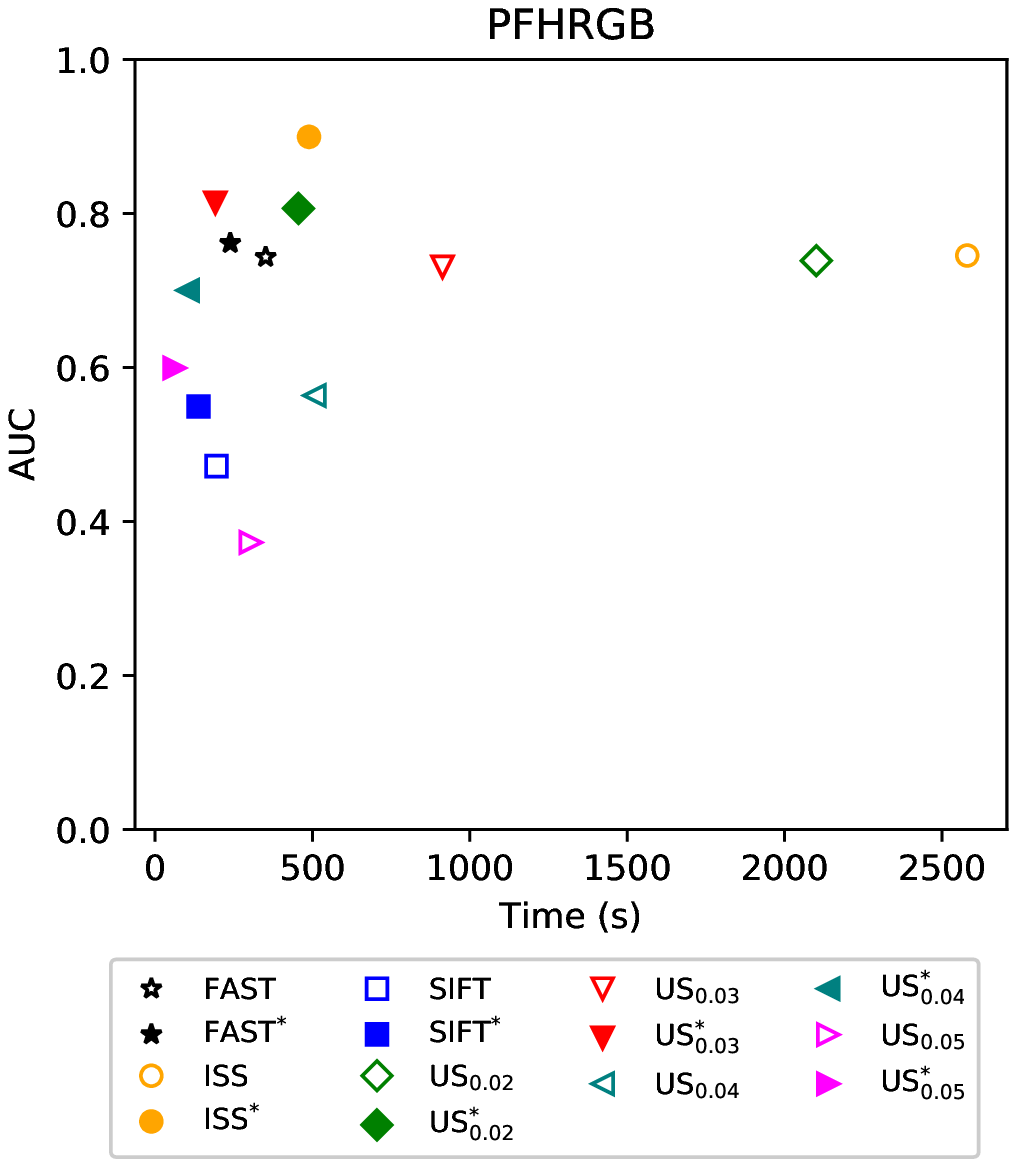}
\label{fig:pfhrgb-results}
}
\subfigure{
\includegraphics[width=0.47\textwidth]{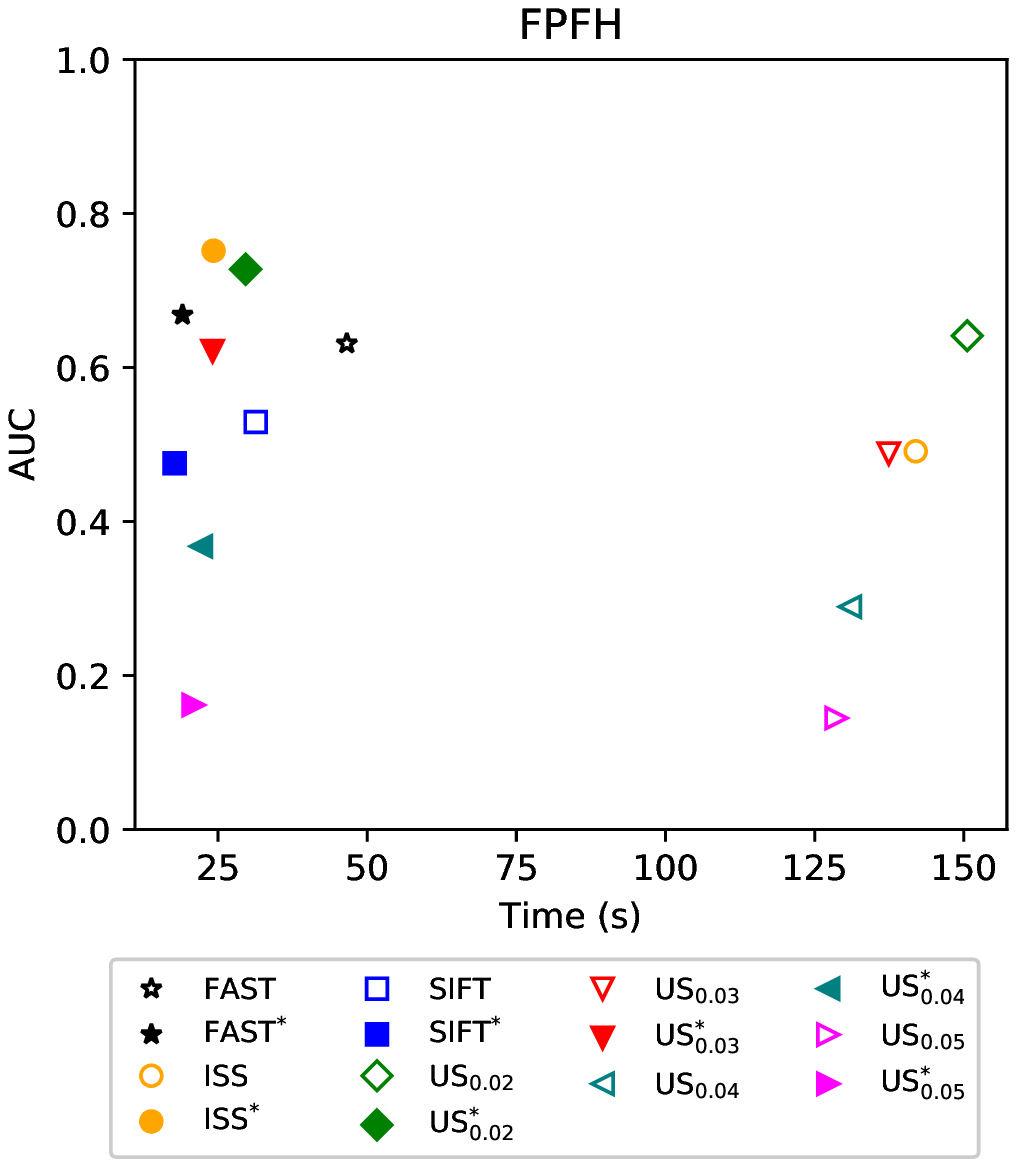}
\label{fig:fpfh-results}
}
\caption{AUC $\times$ Time Results for the descriptors: \subref{fig:cshot-results} CSHOT,
\subref{fig:shot-results} SHOT,  \subref{fig:pfhrgb-results} PFHRGB and  \subref{fig:fpfh-results} FPFH. Boosted pipeline denoted by an asterisk (*) next of the keypoint name.}
\label{fig:AUC-time-results}
\end{figure}

Finally, in Figure \ref{fig:AUC-time-results}, we report a Pareto analysis on the data for all descriptors. We can see how points (i.e. detector/descriptor pairs) closer to the ideal point (that is $AUC = 1$ and Time as low as possible) are obtained by the execution of the boosted pipeline. In this analysis, the CSHOT, SHOT and FPFH obtained the best performance when paired with the boosted ISS (ISS$^{*}$), while PFHRGB when paired with the Boosted Uniform Sampling at $r = 3 cm$ (US$_{0.03}^{*}$). Hence, the boosting pipeline outperforms the traditional one for all tested descriptors when taking into account the combined effect of processing time and recognition performance.

\section{Conclusion}

In this work, we presented an approach based on saliency detection to boost the processing time of the traditional local descriptor pipeline. It was verified for all the tested cases a significant processing time reduction, from 22 to 83\%. Interestingly, the processing time reduction didn't generally decrease the object recognition performance, as measured by the AUC of the precision recall curves. Actually, an improvement on the performance recognition was found for all descriptors in at least one pairing, up to 5\% for SHOT and CSHOT, and more than 50\% for FPFH and PFHRGB. 

In spite of the improvements in processing time, the whole processing time is not suitable for real-time applications yet. However, the proposed approach offers a considerable speed-up without impact negatively on recognition performance, which brings us a step closer to create an effective and real-time local feature pipeline for 3D object recognition.


%
%
\bibliographystyle{splncs04}
\bibliography{references}

\end{document}